\documentclass[10pt,twocolumn,letterpaper]{article}

\usepackage[pagenumbers]{cvpr} % To force page numbers, e.g. for an arXiv version

\usepackage{listings}
\usepackage{xcolor}

\newcommand{\bs}{\boldsymbol}

\definecolor{cvprblue}{rgb}{0.21,0.49,0.74}
\definecolor{ddt}{HTML}{38a3a5}

\usepackage{fontawesome5} % 加载新版
\usepackage{amsthm}
\usepackage{listings}
\usepackage{xcolor}
\usepackage{multirow}
\usepackage[pagebackref,breaklinks,colorlinks,allcolors=cvprblue,urlcolor=ddt]{hyperref}
\def\paperID{***} % *** Enter the Paper ID here
\def\confName{CVPR}
\def\confYear{2026}

\title{{\color{ddt}UniDDT}: {\color{ddt}Uni}fying Multimodal Understanding and Generation \\ with {\color{ddt}D}ecoupled {\color{ddt}D}iffusion {\color{ddt}T}ransformer}

\author{
Shuai Wang\textsuperscript{1} \quad  Liang Li\textsuperscript{2} \quad Yang Chen\textsuperscript{1} \quad  Ruopeng Gao\textsuperscript{1} \quad  Yao Teng\textsuperscript{3} \quad Limin Wang \textsuperscript{1, {\color{lightgray} \faEnvelope}} \\
$^1$Nanjing University \quad  $^2$ByteDance Seed \quad $^3$University of Hong Kong \\  [0.2cm]
{\bf \url{https://github.com/MCG-NJU/UniDDT}} 
}

\newcommand\blfootnote[1]{%	
  \begingroup
  \renewcommand\thefootnote{}\footnote{#1}%
  \addtocounter{footnote}{-1}%
  \endgroup
}

\begin{document}

\twocolumn[{%
\renewcommand\twocolumn[1][]{#1}%
\maketitle
\vspace{-1em}
\includegraphics[width=0.99\linewidth]{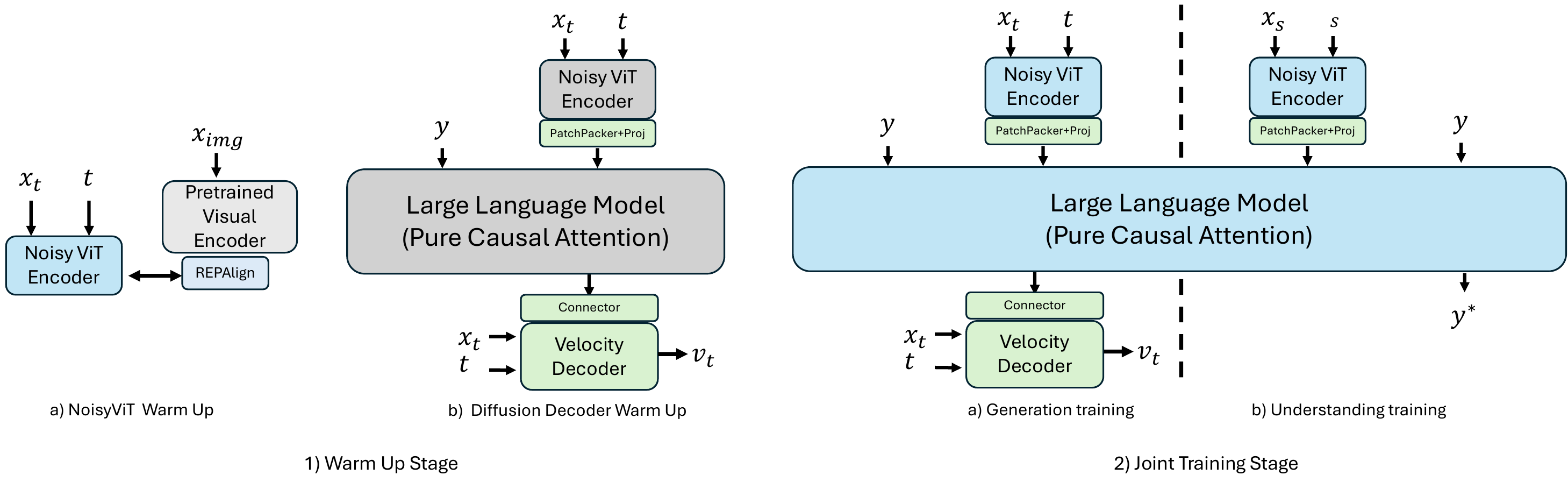}
\vspace{-1em}
\captionof{figure}{\textbf{The two-stages Training Recipe of UniDDT: Warmup training stage and Joint training stage.}~{\small In warmup training, we warm up the Noisy ViT encoder and diffusion decoder separately to avoid model collapse caused by direct joint training. In joint training, we unfreeze all modules and optimize them through a duality of generation and understanding.}}
\label{fig:ddt2_training}
\vspace{1em}
}]

\begin{abstract}
Unified Multimodal Models (UMMs) have emerged as a critical direction for general-purpose multimodal intelligence, integrating understanding and generation into a single framework. However, existing UMMs face prominent challenges: 
(1) the inherent learning conflicts between visual understanding and generation tasks, leading to suboptimal modeling in both tasks;
(2) different understanding and generation visual spaces impeding scalability; 
(3) over-reliance on task-specific data that neglects the duality of text-image understanding and generation. 
To address these challenges, we propose UniDDT, which leverages a Noisy ViT encoder along with an LLM to unify semantic encoding for visual generation and understanding tasks, while employing a separate diffusion decoder to decouple diffusion decoding from text decoding.
With this Noisy ViT encoder, UniDDT is able to leverage the latent space as a unified visual representation, enabling seamless compatibility between understanding and generation tasks. Thus, the scalability within the generation tasks and the semantic expressiveness within understanding tasks can be balanced. Also, we construct dual data structures from the same image-text pairs, fostering interdependence between the generation and understanding data to exploit their inherent duality.
Extensive experiments demonstrate that UniDDT achieves effective unification of multimodal understanding and generation with enhanced semantic consistency and scalability. For visual generation tasks, our UniDDT achieves 0.87 GenEval score and 86.9 DPG overall score. For multimodal understanding tasks, our UniDDT achieves 1699.5 score on MME benchmark and 76.5 overall score on SEEDbench.
\vspace{-2em}
\end{abstract}
\blfootnote{%
{\color{lightgray}\faEnvelope} : 
\begin{tabular}[t]{@{}l@{}}
Corresponding author (lmwang@nju.edu.cn).\\
This work was completed in \textbf{November 2025}.
\end{tabular}%
}
\begin{figure*}
    \centering
    \includegraphics[width=0.98\linewidth]{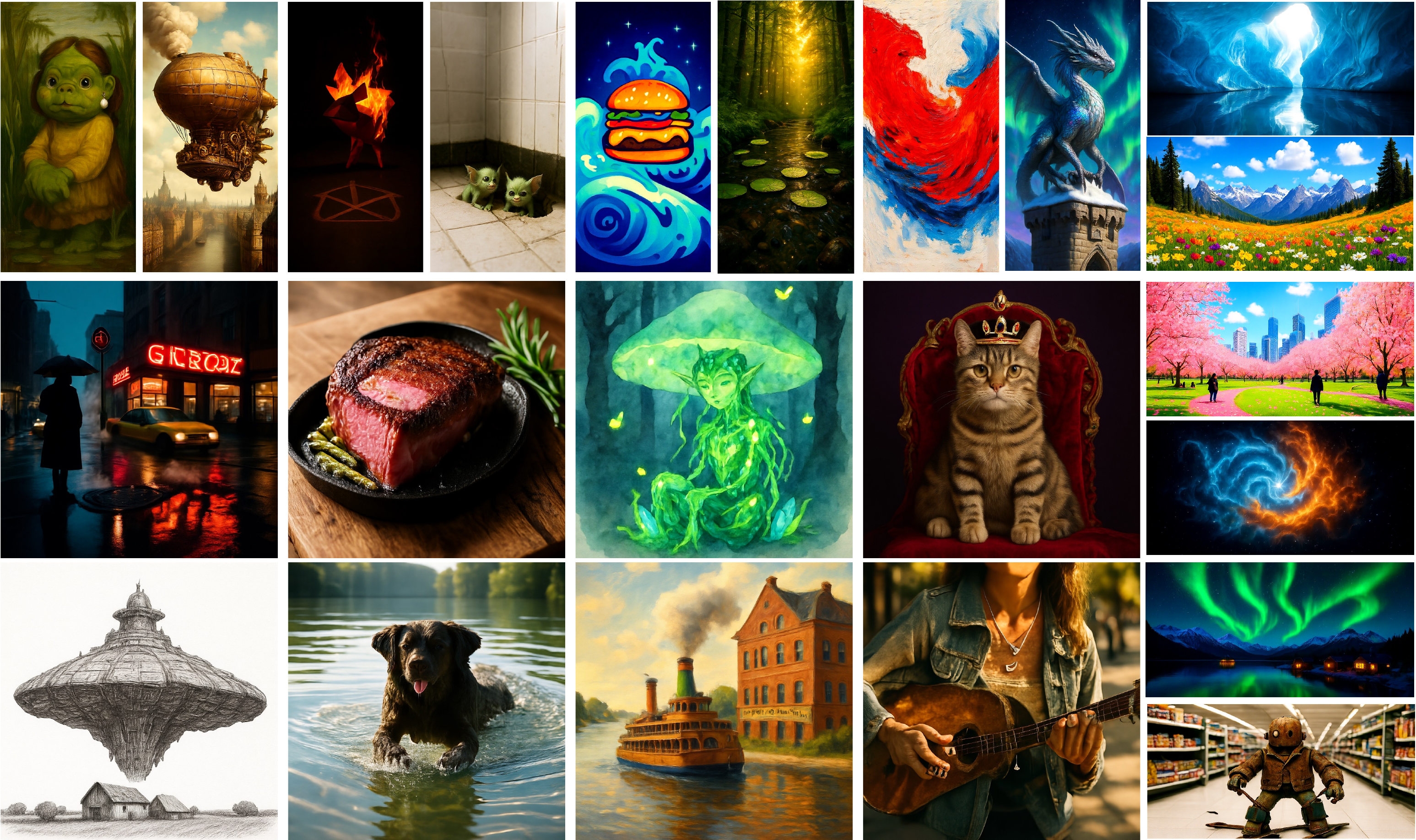}
    \caption{\textbf{The curated samples~(Max Resolution $1024\times1024$) from UniDDT.}~{\small We adopt Adams-2nd solver with 25 steps and CFG value of 4. We place the respective prompts and more visual samples in Appendix.}}
    \label{fig:teaser}
    \vspace{-1.5em}
\end{figure*}
\section{Introduction}

Unified Multimodal Models (UMMs)~\cite{mogao, bagel, metaquery, chameleon, onecat, uniworld, transfusion} integrate both understanding and generation into a single framework. Numerous initiatives~\cite{show_o2, bagel, mogao, onecat} have thrived to close the gap with proprietary unified multimodal systems. Following recent iterative refinement, mainstream text-image UMMs now largely adopt a hybrid AR-diffusion approach (autoregressive for text generation and diffusion for visual generation). However, due to the inherent substantial differences between understanding and visual generation, the implementation of AR-diffusion hybrid UMMs is still neither definitive nor straightforward. We will elaborate on this from three key perspectives: modeling, visual space, and training data.

From Modeling Perspective: UMMs were initially envisioned to achieve mutual promotion between understanding and generation. Early attempts, particularly hybrid AR-diffusion models, used adapters~\cite{metaquery, uniworld} to assemble task-specific tailored models. However, these assembly-based approaches represent a relatively superficial integration, failing to fully exploit the potential synergy between understanding and generation. An alternative approach natively integrates both objectives within a single framework, deemed as Native-UMMs. Mainstream Native-UMMs typically adopt parallel branches for understanding and generation, yet this often results in a significant performance trade-off, still failing to demonstrate the hypothesized mutual promotion. To mitigate this conflict, these UMMs~\cite{mot, bagel, mogao} often decouple the parameters specific to each task.

From the Visual Space Standpoint: A consensus has not yet been reached on the optimal unified visual space. Understanding models thrive on high-dimensional semantic representations, while generative models struggle with training in such spaces. Consequently, most UMMs~\cite{mogao, bagel} adopt distinct visual spaces for different tasks (e.g., a semantic space for understanding and a VAE space for visual generation). This inherent decoupling results in fragmented visual spaces, complicates the overall workflow and impedes large-scale scaling. In response, Unified Visual Space UMMs~\cite{show_o2, onecat} employ a single, shared space. Specifically, some works~\cite{rae} opt for the semantic-rich representations of visual foundation models~\cite{siglip, siglip2, dinov2}, while others\cite{show_o2} choose the detail-rich VAE latent space. Moreover, raw pixel space~\cite{pixnerd, fuyu} seems more scalable, but not validated.

From the Training Data Viewpoint: although Mogao~\cite{mogao} hypothesizes that interleaved multi-modal training data is the key to true unification, most existing UMMs do not follow this approach. Instead, they effectively stitch understanding and generation components together by training each on task-specific data. 
% We hypothesize the unification 
% This is noteworthy because text-image understanding and generation are fundamentally a dual task.

\begin{figure*}
    \centering
    \includegraphics[width=0.95\linewidth]{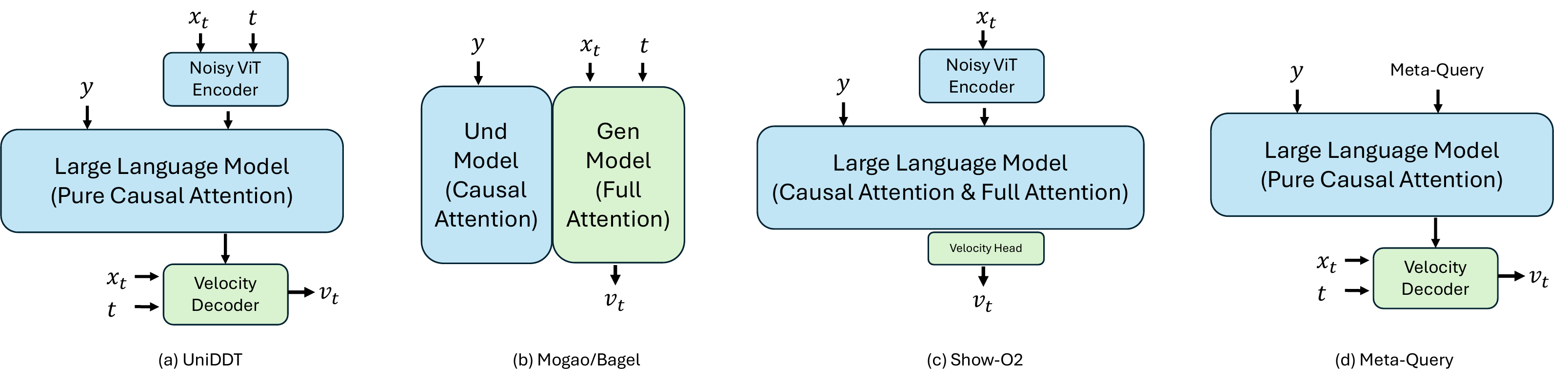}
    \vspace{-1em}
    \caption{\textbf{Architecture comparison with UniDDT and other unified multi-modal models.}~{\small UniDDT consists of a Noisy ViT encoder, an llm backbone, and a diffusion decoder. The Noisy Vit encoder and LLM backbone unify the semantic perception of understanding and generation. The diffusion decoder is dedicated to visual generation.}}
    \label{fig:arch_compare}
    \vspace{-1em}
\end{figure*}

To address the aforementioned problems, we propose UniDDT, a native UMM that features a unified visual space and a decoupled but unified understanding-generation design. An architectural comparison between UniDDT and other UMMs is provided in \cref{fig:arch_compare}. UniDDT leverages a Noisy ViT encoder along with an to unify semantic encoding for visual generation and understanding tasks, while employing a separate diffusion decoder to decouple diffusion decoding from text decoding.

As shown in \cref{fig:ddt2_training}, we treat visual understanding as a preceding task for visual generation. This allows the noisy ViT encoder and LLM backbone to unify two key processes in modeling standpoint: the semantic extraction for standard visual understanding and the semantic encoding of the noisy inputs for diffusion-based generation. The separate diffusion decoder is then dedicated to visual generation, conditioned on the semantics encoded by the ViT and LLM backbone. Regarding the visual space, the Noisy ViT encoder enables the unification of visual spaces, thus we compared pixel and latent spaces choices. Although pixel space holds a slight advantage for understanding, it suffers from a significant deficit in generative performance and does not demonstrate superior scaling properties. Therefore, we adopt the latent space as our principal visual space. From the training data viewpoint, we leverage the understanding-generation duality to enhance our UniDDT under limited data scale. 

% This contrasts with common approaches that rely on separate, task-specific datasets for understanding and generation.

Our contributions are summarized as follows:
\begin{itemize}
    \item We propose UniDDT, a native UMM with a unified visual space and a decoupled but unified understanding-generation design. 
    % \item Our NativeUniDDT-XL achieves 0.89 overall score on GenEval benchmark and 87.1 score on DPG benchmark, surpassing other UMMs with a significant margin.
    \item Our VLM-UniDDT achieves 0.87 overall score on GenEval benchmark and 86.9 score on DPG benchmark, meanwhile, it achieves 1699.5 perception score on MME Benchmark and 76.5 overall score on SEEDbench.
\end{itemize}
\section{Related Works}
\paragraph{Visual Language Models.} Modern Visual Language Models (VLMs)~\cite{internvl, internvl3, internvideo, qwenvl, qwen25vl, qwen2vl} are built on LLMs and trained under the classic next-token prediction paradigm. These VLMs leverage pre-trained visual encoders~\cite{siglip, siglip2} to align raw pixels with the language embedding space. Early attempts explored raw-pixel inputs~\cite{fuyu}, and causal discrete visual tokens~\cite{blip2, minigpt4}, but yielded inferior performance. Other~\cite{cambrian1, eyes} attempts combined different visual encoders for fine-grained perception.
\vspace{-1em}
\paragraph{Visual Generative Models.} High-performance generative models~\cite{seedream2, seedream3, seedream4} typically rely on latent diffusion models. Modern latent diffusion models comprise a variational autoencoder (VAE)~\cite{vavae, ldm, detok, dcae, dcae15, ravae} and a diffusion model~\cite{sit, dit, flowdcn, uvit, ddt}, trained on a latent space shaped by the VAE. Under the classic latent diffusion setup, researchers have leveraged pre-trained visual foundation models to boost performance. Some works~\cite{repa} adopt visual foundation models to align the intermediate features of the diffusion model, while others~\cite{vavae, ravae} align the VAE's latent space. Additionally, DDT~\cite{ddt} enhances generative capability through decoupling diffusion transformer into a tailored architecture. Instead, More ambitiously, RAE~\cite{rae, raev2, rae_scaling} dispenses the traditional VAE and some\cite{pixnerd, pixelflow} strikes back to pixel space. Current discrete visual generation\cite{vqgan, cosmos, fsq, magvit, llamagen} still performs inferior.
\vspace{-1em}

\paragraph{Unified Multimodal Models.} Inspired by the success of large language models, discrete token-based unified models~\cite{unitok, unitoken, dualtoken, emu3, emu35, show_o, chameleon, janus, januspro, dar} convert pixels into discrete visual tokens and are then trained under a unified next-token-prediction paradigm. To mitigate the loss in generative performance in pure discrete auto-regressive approach, AR-diffusion hybrid approaches~\cite{transfusion, show_o2, mogao, bagel, mot, janusflow, rf} leverage discrete auto-regressive modeling for text generation and diffusion modeling for image generation. Beyond native unified frameworks, an alternative research direction~\cite{blip3o, metaquery} involves integrating specialized large multimodal models with diffusion-based generative models by tuning adapters. Concurrently, Representation-Forcing~\cite{rf} proposes to bridge the representation gap across modalities using semantic autoregressive tokens. RepFusion~\cite{repfusion} proposes a unified architecture similar to ours, but further equips it with a powerful RAE~\cite{rae, raev2, rae_scaling}. However, current open-source unified multimodal models still exhibit a significant gap compared to proprietary systems.

\section{Method}

\subsection{Revisit DDT Architecture}
As shown in \cref{fig:arch_compare}, DDT~\cite{ddt} consists of a heavy condition encoder and a light velocity decoder. The heavy condition encoder takes three inputs, a prompt condition $\bs y$, noisy inputs ${\bs x}_t$, and timestep $t$, to extract the self-condition feature ${\bs z}_t$ through stacked diffusion transformer blocks.
\begin{equation}
    {\bs z}_t = \textbf{Encoder}~({\bs x}_t, t, y).
\end{equation}
DDT adopts the representation alignment technique from REPA~\cite{repa} and aligns the intermediate feature $\mathbf{h}_i$ from the $i$-th layer in the condition encoder with the DINOv2 representation $r_*$.  Consistent to REPA~\cite{repa}, the $h_{\phi}$ is the learnable projection MLP:
\begin{equation}
    \mathcal{L}_{enc} = 1-\cos(r_*, h_{\phi}(\mathbf{h_i})) .
    \label{eq:encoder_loss}
\end{equation}
The velocity decoder mirrors the encoder model architecture; it takes the noisy latent ${\bs x}_t$, timestep $t$, and self-conditioning ${\bs z}_t$ as inputs to estimate the velocity ${\bs v}_t$. The external-condition timestep $t$ and self-condition feature ${\bs z}_t$ are used as conditions for the decoder blocks:
\begin{equation}
    {\bs v}_t = \textbf{Decoder}~({\bs x}_t, t, {\bs z}_t).
\end{equation}
The velocity decoder is trained with the flow matching loss as shown in \cref{eq:fm_loss}:
\begin{equation}
    \mathcal{L}_{dec} = \mathbb{E} [\int_0^1 ||({\bs x}_{data} - {\epsilon}) - {\bs v}_t({\bs x_t}, t, {\bs z_t} | \theta)||^2 \mathrm{d}t] .
    \label{eq:fm_loss}
\end{equation}
Finally, DDT jointly trains the condition encoder and the velocity decoder with \cref{eq:encoder_loss} and \cref{eq:fm_loss}. So far, the DDT-like arch has been adopted in RAE~\cite{rae} and PixNerd~\cite{pixnerd}.

\subsection{UniDDT Architecture}
UniDDT adheres to the core philosophy of DDT and is tailored for the \textbf{unified understanding and generation} of text and images. Specifically, UniDDT comprises three key components: a Noisy ViT encoder, a large language model (LLM) backbone, and a diffusion decoder.

As shown in \cref{fig:ddt2_training}, the Noisy Vit encoder takes the noisy input $\bs{x}_t$ and  timestep $t$ as inputs to extract high-level semantics $\bs{s}_t$. If we take the pixel space as the unified visual space, $\bs x_t$ is the noisy image. If we take the latent space as the unified visual space, $\bs x_t$ refers to the noisy latent. The details about unified visual space can be found in \cref{sec:visual_space}. For multimodal understanding, the LLM backbone first performs causal encoding on $\bs{s}_t$, then autoregressively decodes $\bs{y}$. For visual generation, the LLM backbone causally processes the prompt condition $\bs{y}$ and visual semantics $\bs{s}_t$, injecting the semantics from $\bs{y}$ into $\mathbf{\hat{s}}_t$. The diffusion decoder takes the refined visual semantics $\mathbf{\hat{s}}_t$ (derived from $\bs{s}_t$) as the condition, then estimates the velocity $\bs{v}_t$ from the noisy input $\bs{x}_t$. Below, we elaborate on each component in detail:

\paragraph{Noisy ViT Encoder.} Our Noisy Vit encoder mirrors the architecture design as the condition encoder of DDT\cite{ddt}. It is built with interleaved Attention and FFN blocks. The encoder processes two inputs, the noisy latent ${\bs x}_t$ and timestep $t$ to extract the semantic feature ${\bs z}_t$ through a series of stacked Attention and FFN blocks:
\begin{equation}
    {\bs z}_t = \textbf{NoisyViT}~({\bs x}_t, t).
\end{equation}
Similar to DiT~\cite{dit} and SiT~\cite{sit}, we inject the timestep condition through AdaLN-zero~\cite{dit}.
\paragraph{LLM Backbone.} Following the common practice of Qwen-VL~\cite{qwenvl}, we construct distinct chat templates for understanding and visual generation tasks, replacing the image placeholder token with the corresponding image semantic features $\bs{z}_t$ extracted by the Noisy ViT. For multimodal understanding, the LLM backbone first causally encodes the visual semantics $\bs{z}_t$ and understanding prefix tokens (denoted as $\bs{y}$), then autoregressively decodes new text tokens ${\bs y}^*$:
\begin{equation}
    {\bs y}^* = \text{LLM}\left( \bs{z}_t, \bs{y} \right)
\end{equation}
For the visual generation task, the LLM backbone causally encodes the prompt condition $\bs{y}$ and visual semantics $\bs{s}_t$ to perform semantic injection, yielding refined $\hat{\bs{z}}_t$. These refined visual features $\hat{\bs{z}}_t$ are then fed into the diffusion decoder:
\begin{equation}
    \hat{\bs{z}}_t = \text{LLM}\left( \bs{y}, \bs{z}_t \right)
\end{equation}

\paragraph{Diffusion Decoder.} It adopts the same architectural framework as the Noisy ViT encoder, comprising stacked interleaved Attention and FFN blocks—similar to DiT/SiT. It takes the noisy latent $\bs{x}_t$, timestep $t$ as inputs, and $\hat{\bs{z}}_t$ as a condition to estimate the velocity $\bs{v}_t$. Through experiments, we found that training the diffusion decoder using only $\hat{\bs{z}}_t$ (without text tokens) is feasible, even when the LLM backbone and Noisy ViT encoder are frozen. Unlike DDT~\cite{ddt}, we use attention (instead of AdaLN-zero) to inject the $\hat{\bs{z}}_t$ condition into the diffusion decoder features. As shown in \cref{eq:block}, we elaborate on the diffusion decoder's block structure. For readability, we retain the notation $\mathbf{x}_t$ for the intermediate feature:
\begin{align}
    \label{eq:block}
    \mathbf{x}_t &= \mathbf{x}_t + \textsc{AdaLN}\left( \bs{t}, \textsc{Attention}\left( \mathbf{x}_t,\mathbf{\hat{\bs z}}_t \right) \right), \\
    \mathbf{x}_t &= \mathbf{x}_t + \textsc{AdaLN}\left( \bs{t}, \textsc{FFN}\left( \mathbf{x}_t \right) \right).
\end{align}
To improve the training stability, we add several full attention transformer blocks as the refiner\cite{pixnerd, fluid} to refine the provided last hidden states.

\subsection{Unified Visual Space} Previous understanding model, typically, VLMs~\cite{qwenvl, qwen25vl} takes the raw pixels as the principal visual space. While generative models rely on a largely compressed latent space~\cite{vavae, dcae, ravae, ldm} to eliminate redundancy and ease generative model learning. Thus, there exists a learning space tradeoff for a unified model. We found that the understanding performance of the latent space is marginally inferior to that of the pixel space, the performance degradation is minimal, so they can be regarded as comparable in terms of understanding. When it comes to generation performance, though, the latent space is significantly better than the pixel space, and no better scaling advantage has been identified for the pixel space compared to the latent space in our experiments. This motivates us to take the latent space as the principal visual space of UniDDT. \label{sec:visual_space}

\subsection{UniDDT Training}
\paragraph{Warmup Training.} Starting joint training from random initialization can easily cause language model collapse; thus, we employ a separate warmup stage. We use a pre-trained vision-language model (VFM), e.g., SigLIP \cite{siglip, siglip2} or Qwen3-ViT~\cite{qwen3}, as the teacher model to distill representations to the Noisy ViT encoder. Once the Noisy ViT encoder converges, we freeze its parameters along with those of the LLM backbone, then warm up the diffusion decoder.

Specifically, for the Noisy ViT encoder, we initialize most of its parameters from the teacher model, except for the timestep AdaLN-Zero modules. Note that different from show-o2~\cite{show_o2}, our Noisy ViT encoder also takes $t$ as an extra condition for semantic extraction. We then distill representations from the teacher model to the Noisy ViT encoder as defined in \cref{eq:encoder_loss}.For the diffusion decoder warmup (as shown in \cref{fig:ddt2_training}), we add a projection layer to align the dimensions of the Noisy ViT and LLM backbone if needed. We freeze the Noisy ViT encoder and LLM backbone, and jointly train this projection layer and the diffusion decoder using the flow-matching loss specified in \cref{eq:fm_loss}.

\paragraph{Joint Training.} Previous unified models used distinct image-text pairs for understanding and generation tasks. We argue that understanding and generation can be framed as a dual task and leverage this duality to boost our UniDDT under limited data scale.
\label{sec:joint_training}
Specifically, given a text-image pair $(\mathbf{y}, \mathbf{x})$, we construct the data formats as follows (the actual template is provided in the Appendix):
\begin{quote}
\texttt{<user>generate.${\bs y}$<user><bot>${\bs x}$<bot>}
\end{quote}

The understanding-oriented data is constructed as:
\begin{quote}
\texttt{<user>describe.${\bs x}$<user><bot>${\bs y}$<bot>}
\end{quote}

During this stage, we unfreeze all modules to initiate training: we randomly sample between the understanding and generation formats, apply only the cross-entropy loss to the text $\bs{y}$ in the understanding task, and use the diffusion loss for $\bs{x}$ in the generation task. Through experiments, we find this joint training stage benefits visual generation a lot. The joint loss is defined as:
\begin{equation}
    \mathcal{L}_\textit{joint} = \mathbb{E}_{\text{gen}} \mathcal{L}_\textit{diff}(\bs{x}|\bs{y}) + \lambda_{\text{und}} \mathbb{E}_{\text{und}} \mathcal{L}_{\text{ce}} (\bs{y}|\bs{x})
\end{equation}

\paragraph{Post Training.} After the joint training stage, UniDDT can not only generate novel images but also understand the intermediate states of the generation process. This inspires us to further enhance generation quality by leveraging this unique property. Specifically, we freeze the Noisy ViT encoder and LLM backbone, and only train the diffusion decoder during the post-training stage. Given a noisy input $\bs{x}_t$, its corresponding timestep $t$, and the prompt $\bs{y}$, UniDDT takes these three inputs and yields the estimated velocity $\bs{v}_t$. A noisy point at time $s$ can be estimated as: $\bs{x}_s = \bs{x}_t + \bs{v}_t (s - t)$. 
\begin{equation}
    \bs{x}_s = \bs{x}_t + \bs{v}_t({\bs x_t}, t | \theta) (s - t)
\end{equation}
We then feed the intermediate states $\{\bs{x}_s, s\}$ into UniDDT's understanding branch to estimate the likelihood $\log p(\bs{y} | \bs{x}_s, s)$, and maximize this likelihood to improve generation quality and semantic consistency:
\begin{equation}
    \mathcal{L}_\textit{post} = \mathbb{E}_{\bs{x}, t, s, \bs{y}} \mathcal{L}_{\text{ce}}(\bs{y} | \bs{x}_s, s)
    \vspace{-2em}
\end{equation}
\begin{figure}
    \centering
    \includegraphics[width=0.98\linewidth]{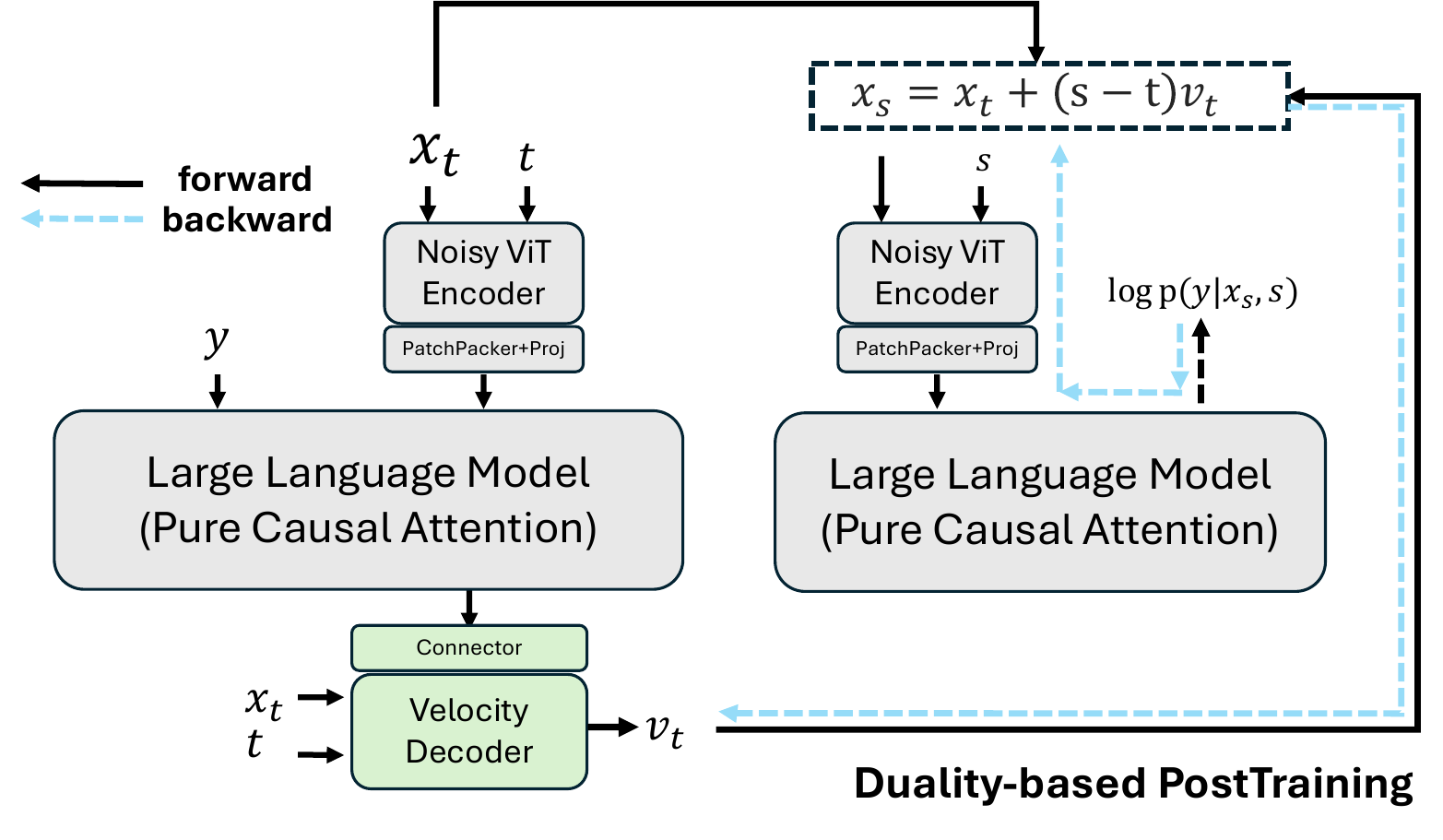}
    \caption{\textbf{The duality-based post-training of UniDDT.}~{\small We freeze the understanding related components and only unfreeze the diffusion decoder; then we feed the intermediate results of visual generation to understanding branch to maximize the likelihood.}}
    \label{fig:post_training}
    \vspace{-1em}
\end{figure}
% Recently, UAE\cite{uae} also adopts CLIP-based reward and GRPO\cite{deepseekmath} to boost the visual generation of UMMS with the understanding capability from the UMM itself.
\section{Experiments}
\paragraph{Dataset.} We collect a mixed dataset with approximately 70M images from publicly available datasets\cite{JourneyDB, midjounery23m, megalith, imagenet, cc12m, commoncatalog}. We recaption every image with Qwen2.5-VL-7B~\cite{qwen25vl} to yield captions with various lengths. The data sources and caption details can be found in the appendix. We place the detailed chat templates for generation and understanding in the Appendix. 

\paragraph{Model Details.}We provide detailed model specifications in \cref{tab:model_specs}. We instantiate two UniDDT variants, differing in their backbone architectures: one with an LLM backbone (denoted as NativeUniDDT) and the other with a VLM backbone (denoted as VLMUniDDT). As shown in \cref{tab:model_specs}, NativeUniDDT-B is configured with a 12-layer Noisy ViT encoder, Qwen3-0.6B\footnote{\url{https://huggingface.co/Qwen/Qwen3-0.6B}} as its LLM backbone, and a 20-layer diffusion decoder with 1024 dimensions(4-layer refiner). NativeUniDDT-L features a 24-layer, 1024-dimension Noisy ViT encoder, a 20-layer, 1536-dimension diffusion decoder, and adopts Qwen3-1.7B\footnote{\url{https://huggingface.co/Qwen/Qwen3-1.7B}} as its LLM backbone. For NativeUniDDT-XL, we scale the diffusion decoder dimension of NativeUniDDT-L to 2560. VLM-UniDDT adopts Qwen3-VL-4B\footnote{\url{https://huggingface.co/Qwen/Qwen3-VL-4B}} as the LLM backbone, while other components are configured consistently with NativeUniDDT-L.  We adopt the latent space of Flux-VAE\footnote{\url{https://huggingface.co/diffusers/FLUX.1-vae}} as the unified visual space of UniDDT.
\begin{table}[h]
    \centering
    \small
    \resizebox{\linewidth}{!}{ % 缩放表格至文本宽度，高度按比例自适应
    \begin{tabular}{l c c c c c c c}
    \toprule
     \multirow{2}{*}{\textbf{Config}} & \multirow{2}{*}{\textbf{LLM}} & \multicolumn{2}{c}{\textbf{NoisyViT}}& \multicolumn{3}{c}{\textbf{Diffusion Decoder}} \\
     & & \#Layers & \#Dim & \#Layers & \#Heads & \#Dim \\
     \midrule
     NativeUniDDT-B & Qwen3-0.6B & 12 & 768 & 16+4 & 16 & 1024\\
     NativeUniDDT-L & Qwen3-1.7B & 24 & 1024 & 16+4 & 24 & 1536\\
     NativeUniDDT-XL & Qwen3-1.7B & 24 & 1024 & 16+4 & 20 & 2560\\
     \midrule
     VLM-UniDDT & Qwen3VL-4B & 24 & 1024 & 16+4 & 24 & 1536\\
     % VLM-DDT2-XL & Qwen3VL-4B & 24 & 1024 & 16 & 20 & 2560\\
     \bottomrule
    \end{tabular}
    } 
    \label{tab:model_specs}
    \caption{\textbf{The detailed model configurations of UniDDT.} {\small UniDDT has two distinct variant, differing in their LLM backbones. NativeUniDDT adopts Qwen3 as its LLM backbone while VLM-UniDDT uses Qwen3-VL as its llm backbone.}}
    \vspace{-2em}
\end{table}

\paragraph{Training Details.} To avoid the mismatch of training text-image pairs caused by center crops, we adopt native aspect ratio training~\cite{nit}. We adopt FSDP~\cite{pytorch, fsdp} to shard model parameters, eliminating memory redundancy. For the Native-UniDDT, we choose SigLIP2~\cite{siglip2} as the teacher model of the Noisy Vit encoder. SigLIP2-B for NativeUniDDT-B, SigLIP2-so-400M for Native-UniDDT-L and Native-UniDDT-XL, respectively. For VLM-UniDDT, we adopt the original visual encoder(Qwen-NaViT~\cite{qwen3}) as the teacher model for the Noisy ViT encoder. In the warmup stage, we train the Noisy ViT encoder for 40K steps with a constant learning rate of 2e-4 and ema rate of 0.9999. After obtained the well-initialized Noisy ViT encoder, we add a proj layer to align the Noisy ViT dimension to the LLM backbone, and jointly train the diffusion decoder and the aforementioned proj layer for 100K steps at maximal sequence length of 16384. For the joint training stage, we unfreeze all modules and optimize with a maximal sequence of 8192~(120K steps for Native-UniDDT, 10K steps for VLM-UniDDT). In order to further enhance the performance, we follow the common practice~\cite{blip3o} to finetuning our UniDDT on OpenAI-4o datasets~\cite{blip3o, echo4o, open4o} for 8K steps.
Our default training hardware consists of $16\times$ A100.

\subsection{Visual Space}
We adopt the latent space of Flux-VAE and the raw pixel space as the candidate visual spaces. The latent space of Flux-VAE has 16 channels with a down-sample factor of 8. Our findings reveal that the pixel space exhibits a slight advantage over the latent space in understanding, though this is accompanied by poorer scaling properties in visual generation during the pretraining stage.
\paragraph{Understanding Perspective.} We collected understanding metrics under the VLM-UniDDT setting, which enabled us to readily obtain cosine similarity and multimodal understanding metrics. The cosine similarity reflects, to some extent, the potential for multimodal understanding. As shown in \cref{fig:nvit_cos_sim}, we fed clean images into the teacher model and noisy images with varying noise timesteps into the Noisy ViT encoder, then calculated the similarity between their features. The Noisy ViT in pixel space exhibited slightly better and more stable similarity across different noise levels, though the performance gap was negligible. The cosine similarity for Native-UniDDT followed a similar trend. For multimodal understanding metrics, we replaced the original visual encoder of Qwen3-VL-4B with our Noisy ViT encoder to collect multimodal performance data. As shown in \cref{fig:nvit_mme}, the noisy ViT encoder in pixel space still performed slightly better.
\paragraph{Generation Perspective.} We collected generation metrics under the Native-UniDDT setting. Across all training stages and spaces, visual generation performance exhibited a clear scaling property. In the warmup and joint training stages, show in \cref{fig:joint_training_scaling} and \cref{fig:warmup_scaling}, the pixel space did not demonstrate better scaling properties than the latent space. In the post-training stage, shown in \cref{fig:post_training_scaling}, the pixel space appeared to perform better, yet the performance gap still remained significant.

\subsection{Multimodal Understanding}
\begin{figure*}
    \centering
    \includegraphics[width=0.75\linewidth]{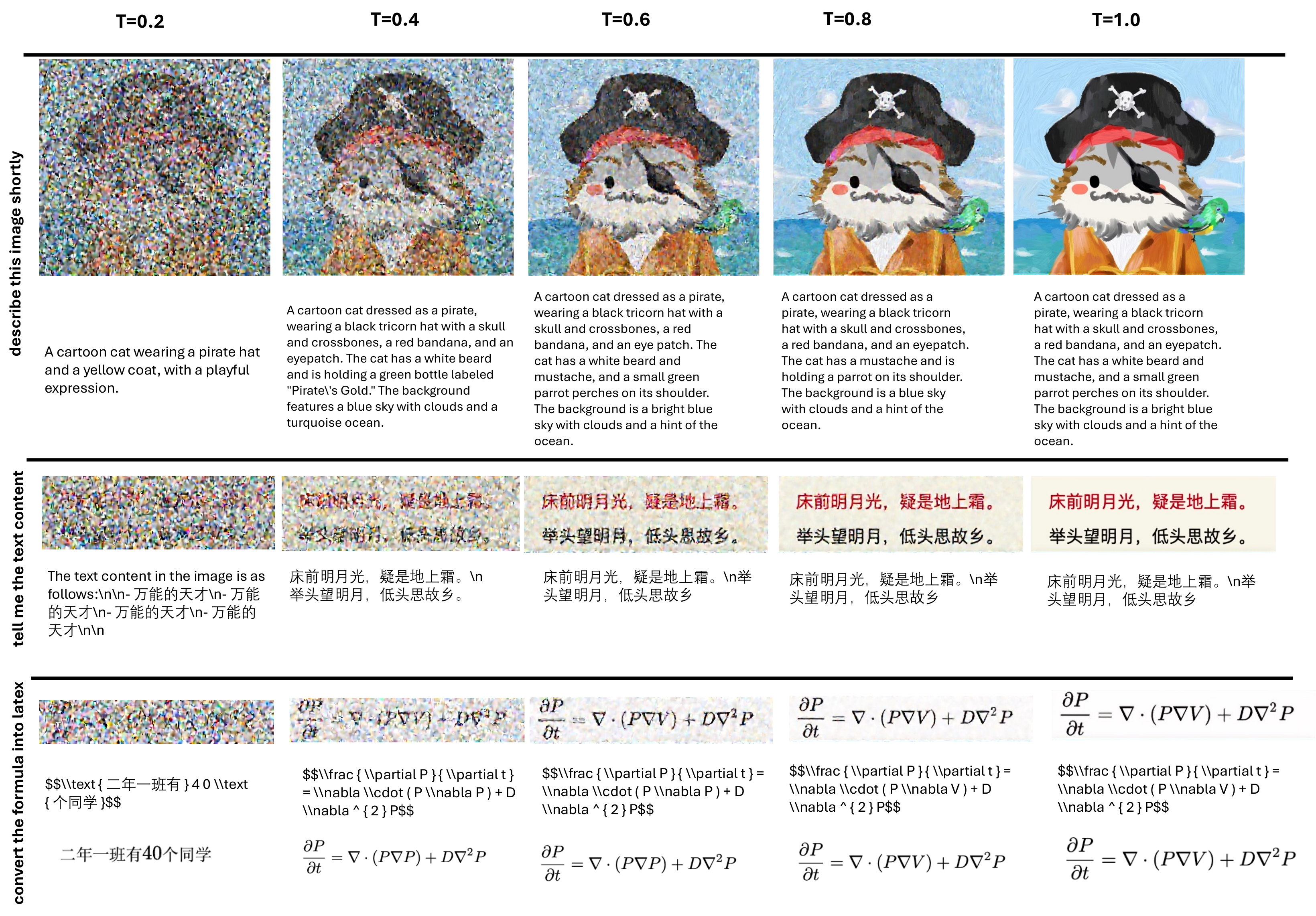}
    \vspace{-1em}
    \caption{\textbf{The understanding power of UniDDT on Noisy inputs.}~{VLM-UniDDT understands inputs well under acceptable noise levels.}}
    \label{fig:und_samples}
\end{figure*}
We fix the timestep $t=1.0$ for multimodal understanding evaluation, but note that the timesteps of the understanding training in joint training stage is randomly sampled from $[0, 1]$. Since our data-source only consists of naive image-text pairs, our Native-UniDDT is only capable of caption images and not follows the instructions, thus we decide not to include the understanding metrics of Native-UniDDT in \cref{tab:und}. We provide the understanding power of UniDDT on Noisy inputs in \cref{fig:und_samples}. We collect the understanding performance on MME~\cite{mme}, SEEDbench~\cite{seedbench}, MMMU~\cite{mmmu}, MMStar~\cite{mmstar}, AI2D~\cite{ai2d} benchmarks.As shown in \cref{tab:und}, our VLM-UniDDT initialized from Qwen3-VL-4B, achieves superior understanding performance across different benchmarks.
\begin{table*}[t] 
\centering
\label{tab:mmu_comparison}
\resizebox{0.8\linewidth}{!}{ 
\begin{tabular}{lcccccccc}
\toprule
{Models} &  \# Params. & MME(p) $\uparrow$ & GQA$ \uparrow$ & SEED$\uparrow$ & MMB(en)$\uparrow$ & MMMU(val) $\uparrow$ & MMStar $\uparrow$ & AI2D $\uparrow$\\
\midrule 
& \multicolumn{6}{c}{\textbf{Understanding. Only}} \\
\midrule
{LLaVA-v1.5~\cite{llava}} & {7B} & {1510.7} & {62.0} & {58.6} & {64.3} & {-} & {-} & {-}\\
{Qwen-VL~\cite{qwenvl}} & {7B} & {1487.6} & {57.5} & {58.2} & {60.6} & {-} & {-} & {57.7} \\
{LLaVA-OV~\cite{llavaonevision}} & {7B} & {1580.0} & {-} & {-} & {80.8} & {48.8} & {57.5} & {81.4} \\
\midrule 
& \multicolumn{6}{c}{\textbf{Unified Models}} \\
\midrule
MetaMorph~\cite{metamorph} & 8B &- & - & 71.8 & 75.2 & - & - & -  \\
{TokenFlow-XL$^{*}$~\cite{tokenflow}} & {14B} & {1551.1} & {62.5} & {72.6} & {76.8} & {43.2} & {-} & {75.9} \\
ILLUME~\cite{illume_plus} & 7B & 1445.3 & - & 72.9 & 75.1 & 38.2 & - & 71.4 \\
{BAGEL~\cite{bagel}} & {14B} & {1687.0} & {-} & {-} & {85.0} & {55.3} & {-} & {-} \\
Show-o~\cite{show_o} & 1.3B & 1097.2 & 58.0 & 51.5 & - & 27.4 & - & - \\
JanusFlow~\cite{janusflow} & 1.5B & 1333.1 & 60.3 & 70.5 & 74.9 & 29.3 & - & - \\
Janus-Pro~\cite{janus} & 1.5B & 1444.0 & 59.3 & 68.3 & 75.5 & 36.3 & - & - \\
Emu3~\cite{emu3} & 8B & - & 60.3 & 68.2 & 58.5 & 31.6 & - & 70.0 \\
VILA-U~\cite{vila} & 7B & 1401.8 & 60.8 & 59.0 & - & - & - & - \\
Liquid~\cite{liquid} & 8B & 1448.0 & 61.1 & - & - & - & - & - \\
Janus-Pro~\cite{janus} & 7B & 1567.1 & 62.0 & 72.1 & 79.2 & 41.0 & - & - \\
Mogao~\cite{mogao} & 7B & 1592.0 & 60.9 & {74.6} & 75.0 & 44.2 & - & - \\
Show-o2\cite{show_o2} & 7B & {1620.5} & {63.1} & 69.8 &  {79.3} & {48.9} & {56.6} & {78.6} \\
% \midrule 
% & \multicolumn{6}{c}{\textbf{Ours}} \\
\textbf{VLM-UniDDT} & \textbf{4B+1B} & \textbf{1699.5} & \textbf{62.3} & \textbf{76.5} & \textbf{82.2} & \textbf{52.6} & \textbf{57.7}& \textbf{78.1}  \\
\bottomrule
\end{tabular}
}
\caption{\textbf{Comparison with others on multimodal understanding benchmarks.}}
\vspace{-1em}
\label{tab:und}
\end{table*}

\subsection{Visual Generation}
\begin{table*}[t]
\centering
\resizebox{0.85\linewidth}{!}{%
\begin{tabular}{lcccccccc}
\toprule
\multirow{2}{*}{{METHOD}} & \multicolumn{7}{c}{{GenEval}} & \multirow{2}{*}{{DPGBench ($\uparrow$)}} \\
& {Single Obj.} & {Two Obj.} & {Counting} & {Colors} & {Position} & {Color Attri.} & {Overall ($\uparrow$)}\\
\midrule
\multicolumn{9}{c}{\textbf{Generation. Only}} \\
\midrule
SDv1.5\cite{ldm} & 0.97 & 0.38 & 0.35 & 0.76 & 0.04 & 0.06 & 0.43 & 63.18 \\
SDv2.1\cite{ldm}  & 0.98 & 0.51 & 0.44 & 0.85 & 0.07 & 0.17 & 0.50 & - \\
SD3-Medium\cite{sd3} & 0.99 & 0.94 & 0.72 & 0.89 & 0.33 & 0.60 & 0.74 & 84.08 \\
SDXL\cite{ldm}  & 0.98 & 0.74 & 0.39 & 0.85 & 0.15 & 0.23 & 0.55 & 74.65  \\
PixArt-$\alpha$\cite{pixart} & 0.98 & 0.50 & 0.44 & 0.80 & 0.08 & 0.07 & 0.48 & 71.11 \\
DALL-E 2\cite{dalle2}  & 0.94 & 0.66 & 0.49 & 0.77 & 0.10 & 0.19 & 0.52 & - \\
DALL-E 3\cite{dalle3}  & 0.96 & 0.87 & 0.47 & 0.83 & 0.43 & 0.45 & 0.67 & 83.50 \\
\midrule
\multicolumn{9}{c}{\textbf{Unified Models}} \\
\midrule
Chameleon\cite{chameleon} & - & - & - & - & - & - & 0.39 & - \\
Show-o\cite{show_o}  & 0.98 & 0.80 & 0.66 & 0.84 & 0.31 & 0.50 & 0.68 &- \\
Show-o2-7B\cite{show_o2} & 1.00 & 0.87 & 0.58 & 0.92 & 0.52 & 0.62 & 0.76$\dagger$ & 86.14 \\
Janus\cite{janus}  & 0.97 & 0.68 & 0.30 & 0.84 & 0.46 & 0.42 & 0.61 & 79.68\\
JanusFlow\cite{janusflow}  & 0.97 & 0.59 & 0.45 & 0.83 & 0.53 & 0.42 & 0.63 & 80.09\\
Janus-Pro-1B\cite{januspro}  & 0.98 & 0.82 & 0.51 & 0.89 & 0.65 & 0.56 & 0.73 & 82.63 \\
Janus-Pro-7B\cite{januspro} & 0.99 & 0.89 & 0.59 & 0.90 & 0.79 & 0.66 & 0.80 & 84.19\\
MetaQuery-B\cite{metaquery} & - & - & - & - & - & - & 0.74$\dagger$ & 80.04 \\
MetaQuery-L\cite{metaquery} & - & - & - & - & - & - & 0.78$\dagger$ & 81.10 \\
MetaQuery-XL\cite{metaquery} & - & - & - & - & - & - & 0.80$\dagger$ & 82.05 \\
BLIP3-o-4B\cite{blip3o} & - & - & - & - & - & - & 0.81 & 79.36 \\
BLIP3-o-8B*\cite{blip3o} & - & - & - & - & - & - & 0.84 & 81.60 \\
BAGEL-7B\cite{bagel} & 0.98 & 0.95 & 0.84 & 0.95 & 0.78 & 0.77 & 0.88$\dagger$ & -\\
% \midrule
% \multicolumn{9}{c}{\textbf{Ours}} \\
% NativeDDT2-L& 0.98 & 0.95 & 0.81 & 0.90 & 0.85 & 0.79  & 0.88 & 86.6 \\
% NativeDDT2-XL& 0.98 & 0.96 & 0.80 & 0.92 & 0.87 & 0.80  & 0.89 & 87.1 \\
\textbf{VLM-UniDDT($512\times512$)}& 0.99 & \textbf{0.93} & 0.71 & 0.92 & \textbf{0.85} & \textbf{0.80} & 0.87 & \textbf{86.9} \\ 
\bottomrule
\end{tabular}%
}
\caption{\textbf{Comparison of various methods on GenEval and DPGBench benchmarks.}~{\small $\dagger$ indicates Generation with prompt rewriting.}}
\label{tab:visual_gen_results}
\vspace{-1.5em}
\end{table*}

As shown in \cref{tab:visual_gen_results}, we report the final visual generation performance after fine-tuning on 4o-like datasets~\cite{blip3o, open4o, echo4o}. Native-UniDDT-L achieves an overall GenEval score of 0.88 and a DPG-Bench score of 86.6, while scaling the model to Native-UniDDT-XL further improves the results to 0.89 on GenEval~\cite{geneval} and 87.1 on DPG-Bench~\cite{dpg}. VLM-UniDDT also delivers strong generation quality, achieving 0.87 on GenEval and 86.9 on DPG-Bench.

These results show that UniDDT is competitive with, and in many cases superior to, both dedicated generative models and existing unified multimodal models. In particular, the strong performance on GenEval suggests that UniDDT preserves robust object-level compositionality, while the high DPG-Bench score indicates favorable prompt-following and semantic alignment. Together, these results demonstrate that decoupling diffusion decoding from text decoding does not compromise generation quality; instead, it enables UniDDT to maintain strong visual synthesis capability while sharing a unified semantic modeling pathway for understanding and generation. Qualitative visual samples are provided in \cref{fig:teaser} and the Appendix.

\subsection{Ablation}
\begin{table}
\vspace{0.5em}
\renewcommand\arraystretch{0.5}
\setlength{\tabcolsep}{1pt}
\small
\begin{tabular}{c | c c c c c c}
\toprule
Stage & {Two Obj.} & {Counting} & {Position} & {Color Attri.} & Overall \\
\midrule
Warmup-training & 0.63 & 0.32 & 0.20 & 0.27 & 0.52\\
Joint-training  & 0.69 & 0.45 & 0.41 & 0.37 & 0.60 \\
Post-training   & 0.84 & 0.56 & 0.50 & 0.61 & 0.72 \\
\midrule
+4o data fine-tuning & 0.93 & 0.71 & 0.85 & 0.80 & 0.87\\
\bottomrule
\end{tabular}
\caption{Performance ablation across different training stages.}
\vspace{-2em}
\label{tab:iso_perf}
\end{table}

\begin{figure*}
    \centering
    \subfloat[CosSim of NoisyViT\label{fig:nvit_cos_sim}]{
    \includegraphics[width=0.24\linewidth]{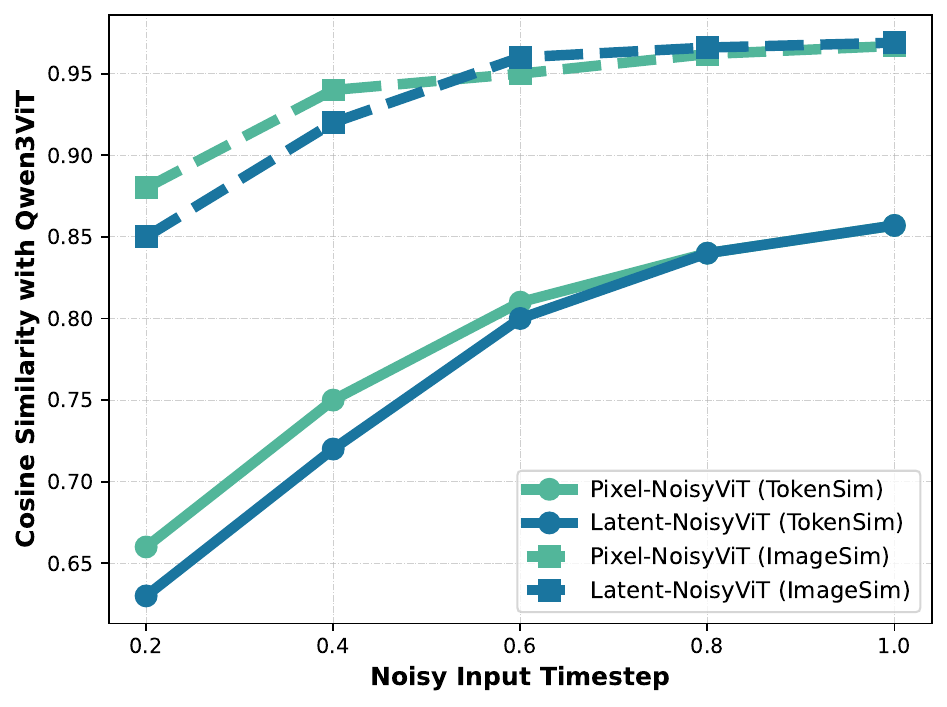}
    }
    \subfloat[Warm Up Stage of Latent Space\label{fig:latent_dit_warmup}]{
    \includegraphics[width=0.24\linewidth]{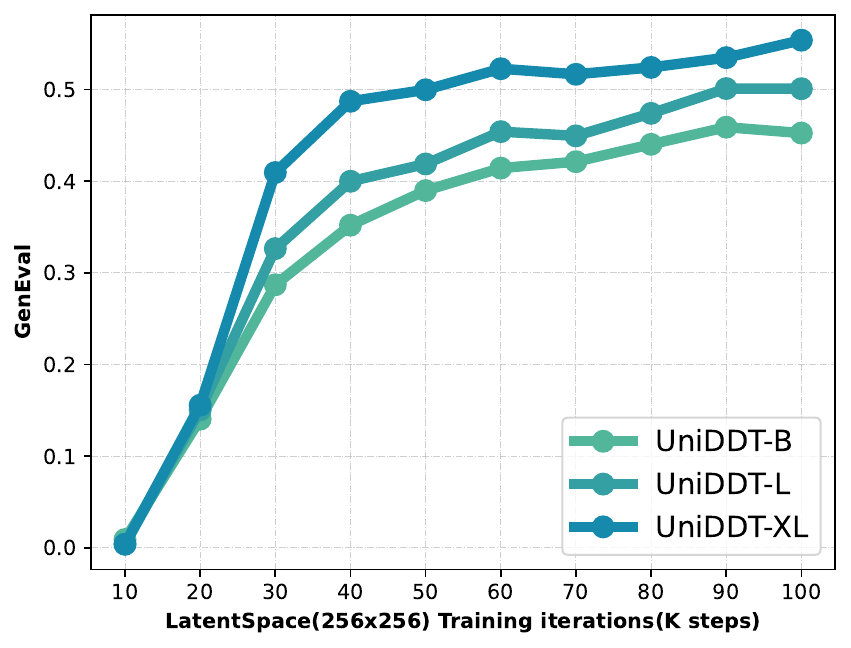}
    }
    \subfloat[Warm Up Stage of Pixel Space\label{fig:pixel_dit_warmup}]{
    \includegraphics[width=0.24\linewidth]{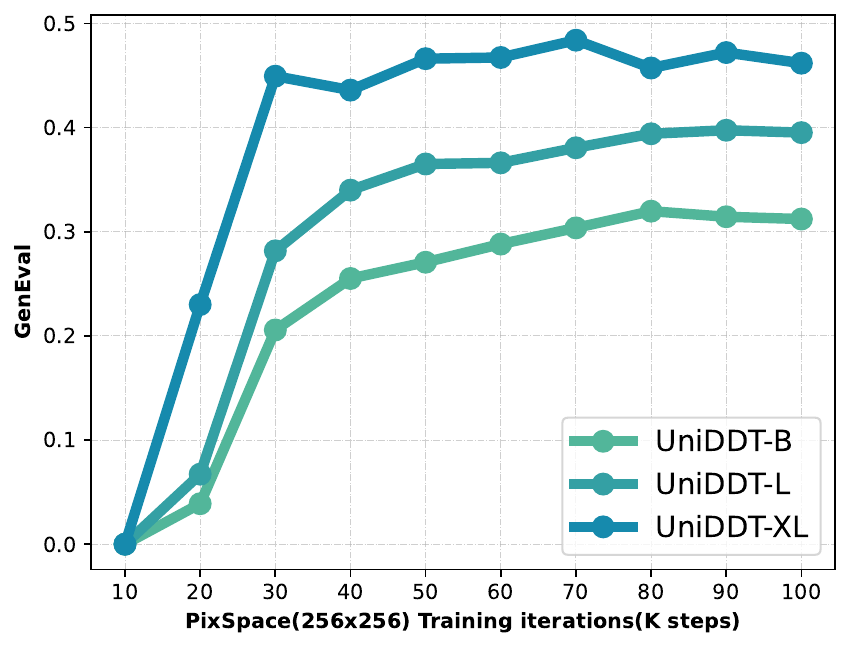}
    }
    % TODO
    \subfloat[Scaling Law of Warmup Training\label{fig:warmup_scaling}]{
    \includegraphics[width=0.24\linewidth]{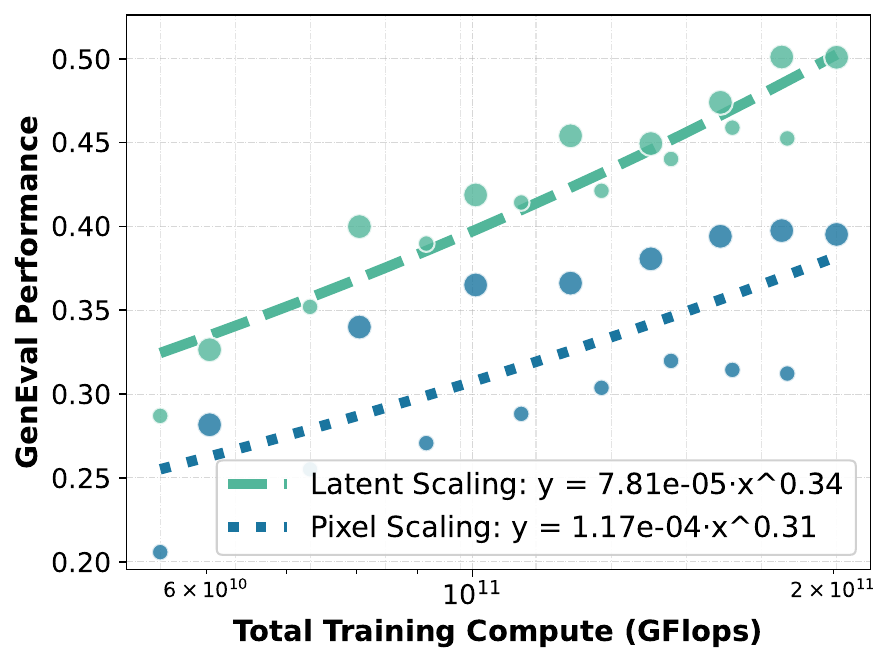}
    }

    \subfloat[MME of NoisyViT\label{fig:nvit_mme}]{
    \includegraphics[width=0.24\linewidth]{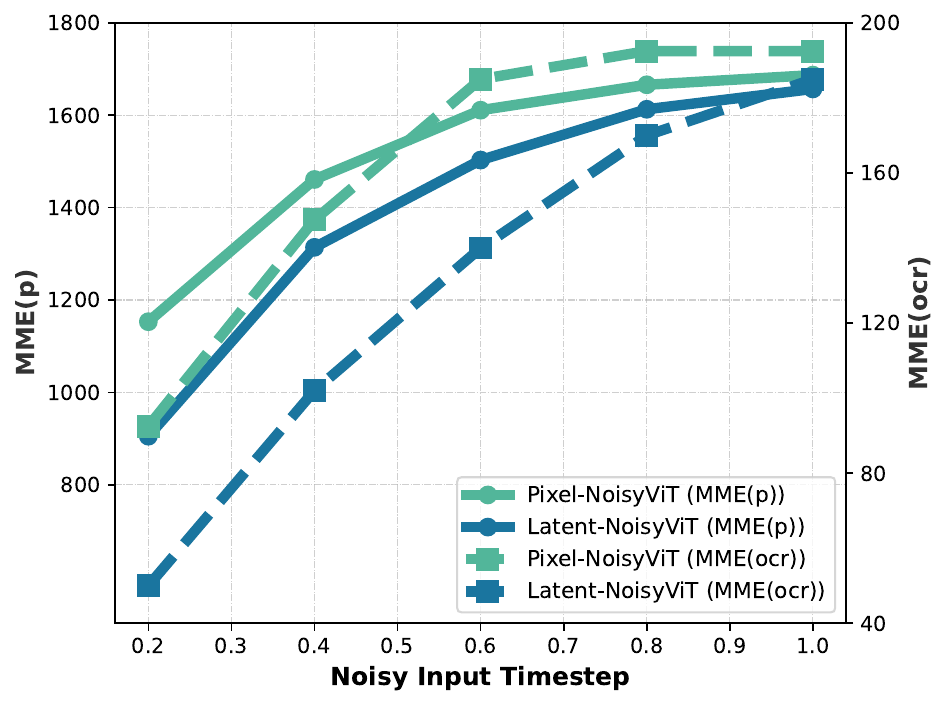}
    }
    \subfloat[Joint Training w/o Understanding\label{fig:joint_training_wo_understanding}]{
    \includegraphics[width=0.24\linewidth]{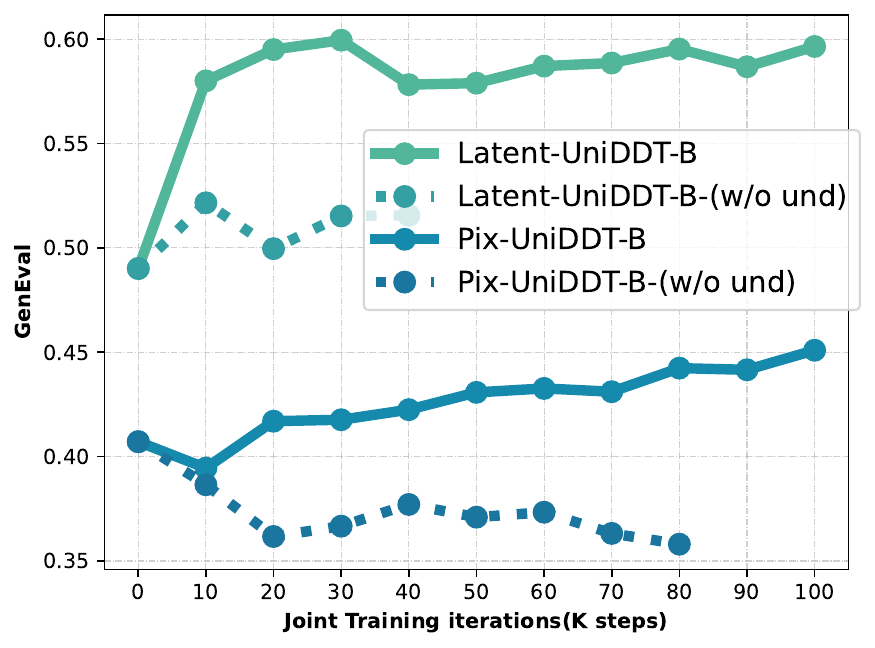}
    }
     \subfloat[Joint Training Dynamics\label{fig:joint_training}]{
    \includegraphics[width=0.24\linewidth]{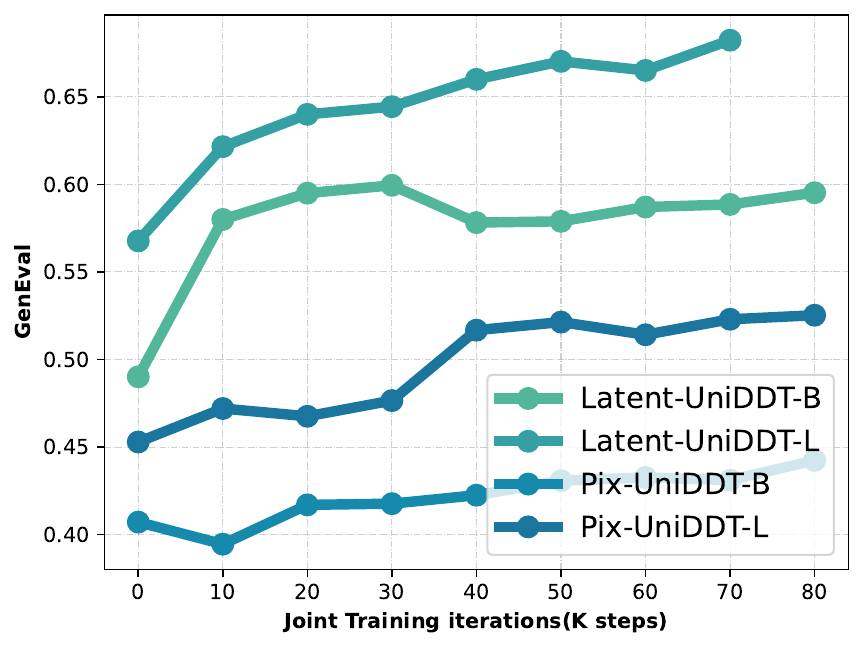}
    }
    % TODO
    \subfloat[The Scaling Law of Joint Training\label{fig:joint_training_scaling}]{
    \includegraphics[width=0.24\linewidth]{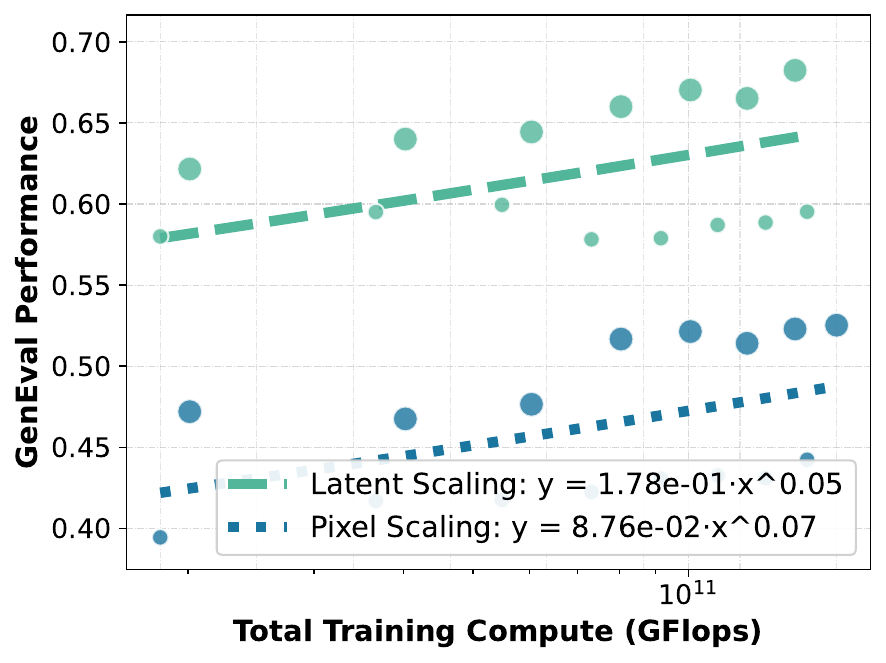}
    }

    \subfloat[Training Timeshift of NoisyViT\label{fig:nvit_mme_timeshift}]{
    \includegraphics[width=0.24\linewidth]{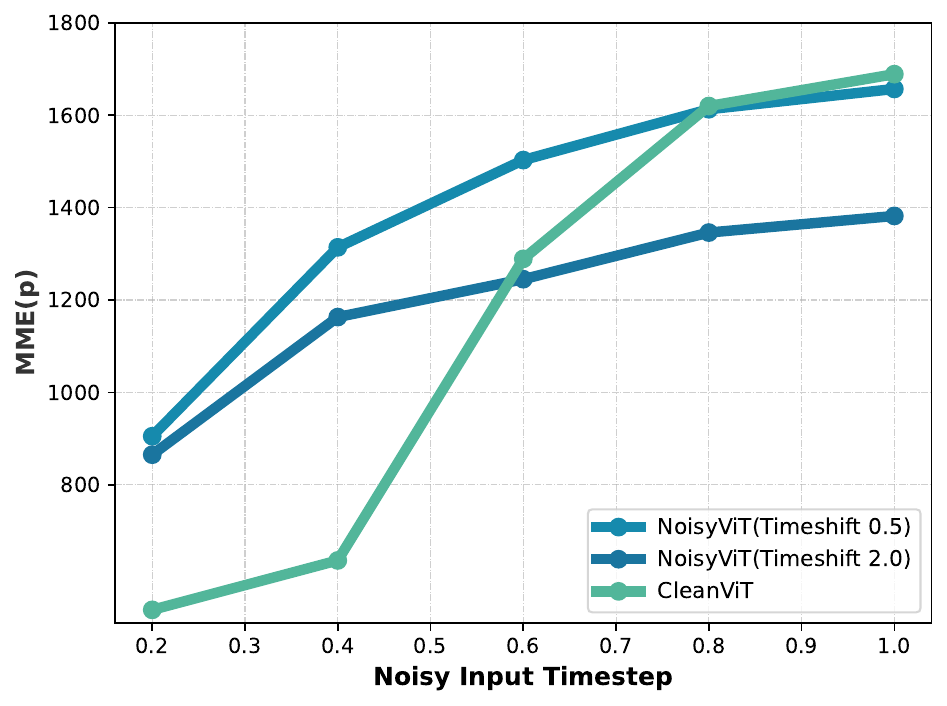}
    }
    \subfloat[Post Training of Latent Space\label{fig:latent_post_training}]{
    \includegraphics[width=0.24\linewidth]{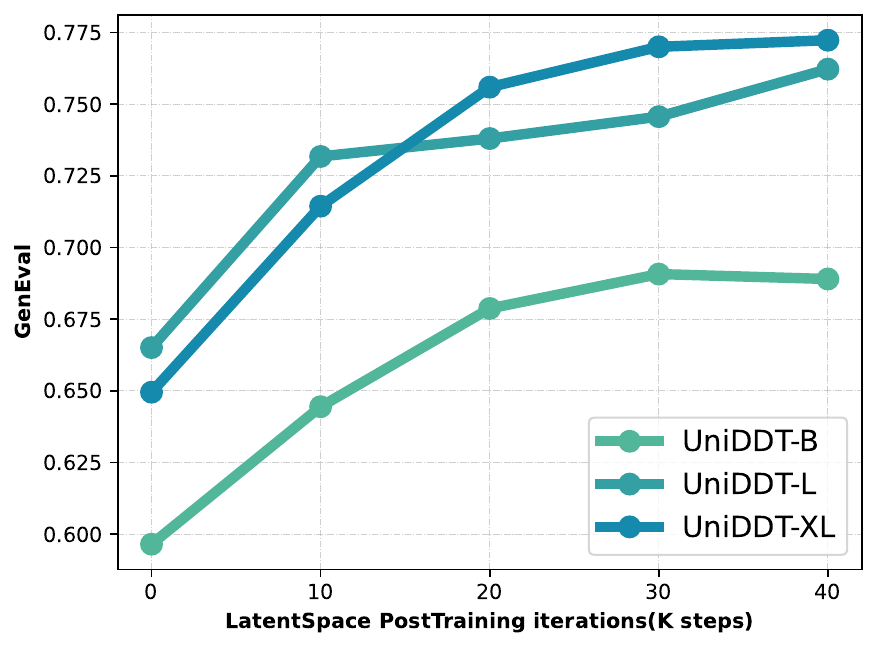}
    }
    \subfloat[Post Training of Pixel Space\label{fig:pixel_post_training}]{
    \includegraphics[width=0.24\linewidth]{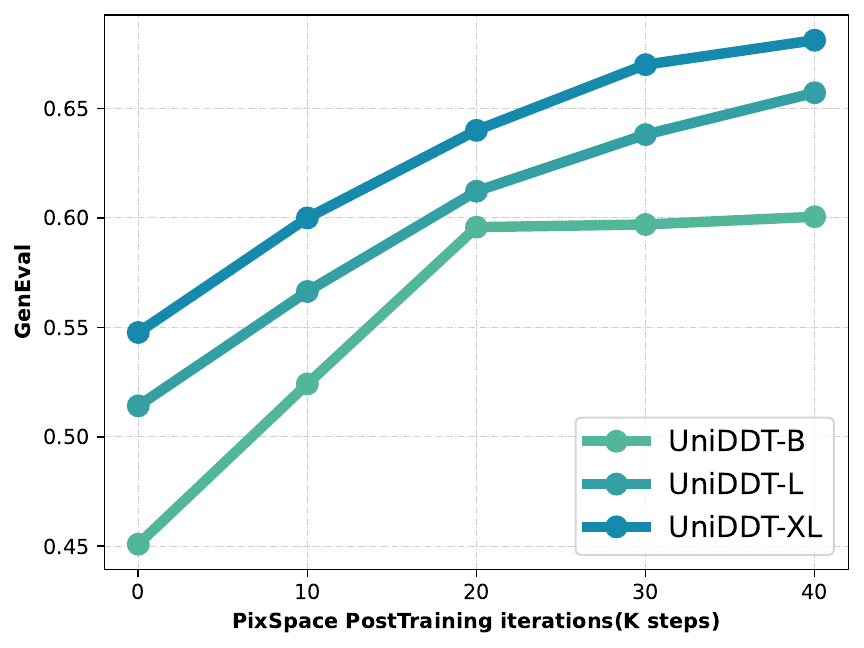}
    }
    % TODO
    \subfloat[The Scaling Law of Post Training\label{fig:post_training_scaling}]{
    \includegraphics[width=0.24\linewidth]{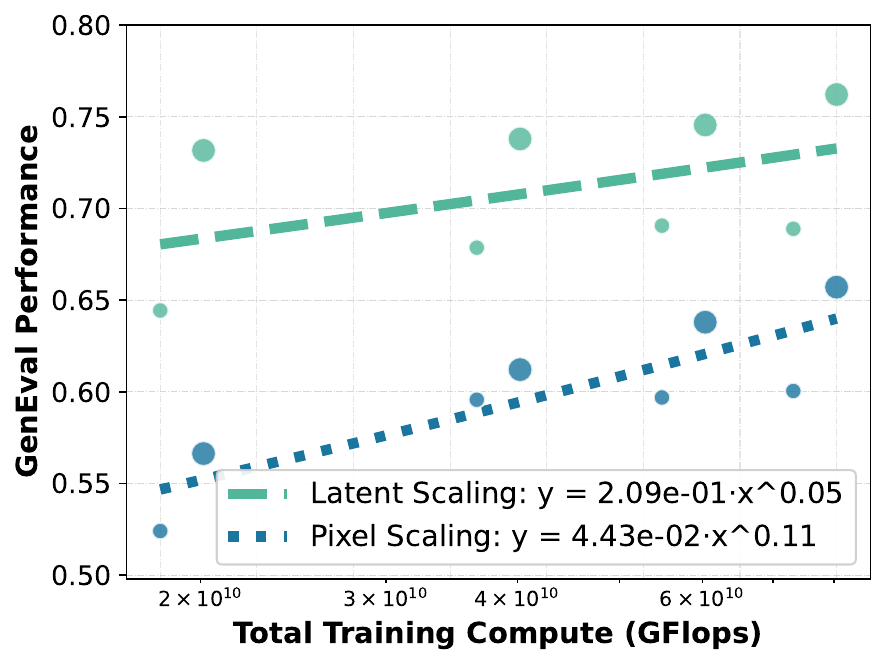}
    }
    \caption{\textbf{The ablation studies table.}~{\small We conduct vigorous ablation experiments on UniDDT. Specifically, we provide the training curves of warmup stage, joint training stage and the post-training stage of UniDDT.}}
    \vspace{-1.5em}
\end{figure*}

\paragraph{Warm-up Noisy ViT Encoder.} Time shift~\cite{sd3} and log-normal~\cite{sd3} play a pivotal role in training diffusion-based generative models. We also adopt these strategies during the warm-up stage of the Noisy ViT encoder. However, as shown in \cref{fig:nvit_mme_timeshift}, we surprisingly found that a commonly used large time-shift value (corresponding to more noisy timesteps) significantly impairs visual understanding performance—particularly OCR capability. This observation inspires us to use a small time-shift value, which generalizes well to more noisy steps. We also include a clean ViT variant, which is trained exclusively on clean inputs. While the clean ViT encoder has limited generalization to noisy timesteps, our Noisy ViT encoder performs consistently well across different noisy timesteps.

\paragraph{Warm-up Diffusion Decoder.} Conventional diffusion transformers use the last hidden states of text tokens from the LLM as conditioning. In contrast, our diffusion decoder exclusively adopts the refined visual features from the LLM backbone. This raises doubts about whether these refined visual features can efficiently summarize the prefix text prompt. As shown in \cref{fig:latent_dit_warmup} and \cref{fig:pixel_dit_warmup}, the diffusion decoder learns effectively even when the Noisy ViT encoder and LLM backbone are frozen. Furthermore, as shown in \cref{fig:warmup_scaling}, performance improves steadily as the allocated training compute increases, no matter in pixel space or latent space.

\paragraph{Joint Training.} Consistent with \cref{sec:joint_training}, we construct joint training data from the same text-image pairs by leveraging the duality between generation and understanding. To validate the effectiveness of the joint training stage for visual generation, we designed a targeted experiment: we unfreeze all modules (consistent with standard joint training) while setting the understanding loss weight to zero. As shown in \cref{fig:joint_training_wo_understanding}, this targeted experiment is denoted as \textit{(w/o und)}. In contrast to the default duality-leveraging joint training, training exclusively on visual generation data yields inferior performance. Specifically, Latent-Native-UniDDT trained in the absence of understanding data achieves only marginal improvements, while Pixel-Native-UniDDT-B even exhibits performance degradation—which further confirms the instability of the pixel space. By contrast, under duality-aware joint training, Latent-Native-UniDDT-B delivers a significant performance leap, and Pixel-Native-UniDDT-B improves steadily. As shown in \cref{fig:joint_training} and \cref{fig:joint_training_scaling}, larger model sizes correspond to superior performance, with joint training exhibiting clear computational scaling behavior.

\paragraph{Post Training.} Empowered by the duality of understanding and generation, our post-training significantly boosts generation performance. As shown in \cref{fig:pixel_post_training} and \cref{fig:latent_post_training}, visual generation performance improves steadily. As illustrated in \cref{fig:post_training_scaling}, duality-based post-training exhibits clear scaling behavior.

\paragraph{Improvements of each stage.} To better isolate the contribution of our architecture from the effects of additional fine-tuning data, we report the performance of VLM-UniDDT at each training stage prior to fine-tuning on OpenAI GPT-4o-style data in the \cref{tab:iso_perf}.

\section{Limitation}
Although we emphasize making full use of the duality of data, the text in the original image-text data mainly comes from captions generated by other models. This greatly limits the understanding ability and instruction following capability of Native-UniDDT leaving it only capable of performing image captioning tasks. Thus, we decide not to report the understanding performance of Native-UniDDT, and only provide the performance of VLM-UniDDT in \cref{tab:und} and \cref{tab:visual_gen_results}. We believe that increasing the richness of the original data helps address this issue. Concurrent work\cite{repfusion} adopting the similar architecture also uses a stronger VAE, which is likewise a promising direction for further exploration and improvement. Since our experiments were conducted before the release of JiT\cite{jit}, our pixel-space experiments did not consider the prediction formulation proposed by JiT, leaving room for further improvement.

\section{Conclusion}
Unified Multimodal Models (UMMs) are pivotal for advancing general-purpose multimodal intelligence, yet existing approaches face core challenges: conflicting objectives between understanding and generation in modeling, fragmented visual spaces that hinder scalability, and task-specific training data failing to leverage text-image duality. To address these, we propose UniDDT, a native UMM with a decoupled yet unified design—comprising a Noisy ViT encoder, an LLM backbone, and a diffusion decoder—that frames understanding as a prerequisite for generation to avoid mutual trade-offs, adopts the latent space as the unified visual representation for balanced semantic expressiveness and scalability, and constructs dual understanding-generation data from the same image-text pairs without relying on task-specific datasets. This concise framework enhances semantic consistency between understanding and generation, demonstrating that decoupling conflicting objectives, unifying visual representation, and leveraging task duality are key to advancing UMMs, and offers a new perspective for next-generation unified model design.

{
    \small
    \bibliographystyle{ieeenat_fullname}
    \bibliography{main}
}

% \appendix
% \newpage
% \input{sec/appendix}

\end{document}